\documentclass[10pt,twocolumn,letterpaper]{article}
\usepackage{times}
\usepackage[pagenumbers]{cvpr}
\usepackage[T1]{fontenc}
\usepackage{graphicx}
\usepackage{booktabs}
\usepackage[misc]{ifsym}

\usepackage{standalone}
\usepackage{hyperref}
\usepackage{multirow}
\usepackage{rotating}
\usepackage[table]{xcolor}
\definecolor{textcolortab}{RGB}{70,170,34}
\usepackage{amsfonts}
\usepackage{amsmath}
\usepackage{amssymb}
\usepackage{mathtools}
\usepackage{svg}
\usepackage{graphicx}
\usepackage{subcaption}
\usepackage[capitalize]{cleveref}
\crefname{section}{Sec.}{Secs.}
\Crefname{section}{Section}{Sections}
\Crefname{table}{Table}{Tables}
\crefname{table}{Tab.}{Tabs.}
\crefname{figure}{Fig.}{Figs.}
\Crefname{figure}{Figure}{Figures}
\usepackage{bbm}
\usepackage{adjustbox}

\begin{document}

\title{Efficient Contrastive Decoding with Probabilistic Hallucination Detection \\
- Mitigating Hallucinations in Large Vision Language Models -}

\author{Laura Fieback$^{1,2}$ \qquad Nishilkumar Balar$^{3}$ \qquad Jakob Spiegelberg$^1$ \qquad Hanno Gottschalk$^2$\\
$^1$Volkswagen AG \qquad $^2$Technical University Berlin \qquad $^3$University of Siegen\\
{\tt\small \{laura.fieback, jakob.spiegelberg\}@volkswagen.de} \\
{\tt\small gottschalk@math.tu-berlin.de} \\
{\tt\small nishilkumar.balar@student.uni-siegen.de}}

\maketitle

\begin{abstract}
Despite recent advances in Large Vision Language Models (LVLMs), these models still suffer from generating hallucinatory responses that do not align with the visual input provided. To mitigate such hallucinations, we introduce Efficient Contrastive Decoding (ECD), a simple method that leverages probabilistic hallucination detection to shift the output distribution towards contextually accurate answers at inference time. By contrasting token probabilities and hallucination scores, ECD subtracts hallucinated concepts from the original distribution, effectively suppressing hallucinations. Notably, our proposed method can be applied to any open-source LVLM and does not require additional LVLM training. We evaluate our method on several benchmark datasets and across different LVLMs. Our experiments show that ECD effectively mitigates hallucinations, outperforming state-of-the-art methods with respect to performance on LVLM benchmarks and computation time.

\end{abstract}

\section{Introduction}
\label{sec:intro}
By aligning textual and visual features, LVLMs have shown an impressive vision-language understanding across various multimodal tasks like visual question answering (VQA) or image captioning \cite{Liu.2023,Dai.2023,zhu2024minigpt}. However, inconsistencies between the generated response and the visual input, a phenomenon called hallucinations \cite{Rohrbach.2018,Li.2023b}, diminish the applicability of LVLMs in safety-critical applications such as autonomous driving \cite{gao2024surveyfoundationmodelsautonomous,tian2024drivevlm} or medicine \cite{Jiang.2024,Li.2023c}. Motivated by recent findings \cite{wang2023haelm,HongYanLijunLiuXupengFeng&QingsongHuang.2022} that identified overreliance of LVLMs on language priors as one of the main reasons for hallucinations, new hallucinatory datasets and fine-tuning strategies have been proposed to mitigate hallucinations \cite{Gunjal.2024,liu2024mitigating,Jiang.2024b}. Contrastive Decoding (CD) strategies \cite{Leng_2024_CVPR,Wang.2024} emerged as a training-free alternative, addressing concerns about computational costs and human effort required for data labeling. The idea of CD is to intervene in the decoding process of LVLMs by amplifying the language prior through distorted inputs and contrasting the output distribution with the distribution derived from original inputs. While this approach effectively mitigates hallucinations and computational overhead, it still increases the inference time by calculating two output distributions.

In this work, we investigate the potential of probabilistic hallucination detection for Efficient Contrastive Decoding (ECD). During the decoding process, token scores are contrasted with hallucination scores to suppress hallucinations. We employ the idea of meta classification \cite{hendrycks2017a,matatoken.2025} to train a lightweight detector to estimate hallucination scores based on hallucination features derived from the model output. By investigating features from intermediate LVLM layers, we achieve area under precision recall curve values \cite{Davis.2006} of up to $74.05 \%$. In contrast to existing CD methods, our approach requires only one forward pass of the LVLM followed by the lightweight hallucination detection, effectively reducing the inference time. Moreover, instead of amplifying hallucinations through input uncertainty, we directly learn hallucinated concepts from internal LVLM calculations, outperforming recent CD methods across several state-of-the-art LVLMs and benchmarks. In detail, ECD mitigates the hallucination rate by up to $5.74 pp$, i.e., $32 \%$ in open-ended tasks and improves F1 Scores by $23.02 pp$, i.e., $33 \%$ in discriminative VQA benchmarks, while adding only minor computational overhead to the decoding process. Our main contributions are as follows:
\begin{itemize}
    \item We propose new hallucination features to train a powerful lightweight hallucination detector.
    \item Based on this detector, we introduce ECD, a lightweight and training-free decoding method that effectively mitigates hallucinations in LVLMs by penalizing mendacious tokens through hallucination scores.
    \item Through extensive experiments, we show the effectiveness of our approach outperforming state-of-the-art methods on various benchmarks and in computational time.
\end{itemize}
\section{Related Work}
\label{sec:related_work}
\subsection{Hallucination Mitigation for LVLMs}
The research area of vision-language pre-trained models has made substantial progress by incorporating Large Language Models (LLMs) building the powerful Large Vision Language Models (LVLMs). In general, LVLMs consist of (i) a vision encoder to extract vision features from the input image, (ii) a cross-modal alignment module, which aligns the visual and language features, and (iii) an LLM, which generates the text response. Despite remarkable zero-shot capabilities in multimodal tasks, LVLMs suffer from hallucinations, i.e., they generate answers that do not align with the input image. Several hallucination mitigation methods have been proposed comprising new instruction tuning datasets for LVLM retraining \cite{liu2024mitigating,Gunjal.2024,Jiang.2024b}, leveraging expert models for post hoc hallucination correction \cite{zhou2024analyzing,woodpecker2024} or incorporating object grounding features \cite{Kim.2025,Chen.2024,LinxiZhaoandYiheDengandWeitongZhangandQuanquanGu.2024}. However, these methods require extensive data collection and annotation, LVLM retraining or architecture changes, which can be time-consuming and computationally costly. To cope with this problem, simple contrastive decoding strategies have been introduced, which contrast the output distributions with original and distorted inputs during inference. Based on the observation that hallucinations often occur due to the overreliance of LVLMs on language priors \cite{Rohrbach.2018}, the authors of Visual Contrastive Decoding (VCD) \cite{Leng_2024_CVPR} proposed to contrast the original output distribution with the distribution derived from noisy input images to subtract the language bias from the original distribution. Similarly, Instruction Contrastive Decoding (ICD) \cite{Wang.2024} adds prefixes to the text input to increase multimodal alignment uncertainty and finally contrasts the resulting distribution with the original output. Although these methods successfully mitigate hallucinations, they increase the inference time by performing one forward pass with original inputs and one with distorted inputs. Instead, we propose to contrast the output distribution with hallucination scores derived from internal LVLM calculations using meta classification, which effectively reduces computational costs during contrastive decoding.

\subsection{Meta Classification for Hallucination Detection} 
In order to judge the reliability of LVLM responses, different hallucination detection methods have been introduced. These methods either apply a pipeline of stacked LLMs and LVLMs \cite{Jing.2024,Wu.2024} to detect hallucinations as a post hoc method or train L(V)LM-based classifiers \cite{wang2023haelm,Gunjal.2024} using hand-crafted hallucination datasets. Since these methods are computationally costly, MetaToken \cite{matatoken.2025} introduced a lightweight and simple hallucination detection method based on meta classification \cite{hendrycks2017a}. In general, meta classification refers to the classification of true and false predictions based on uncertainty features derived from the model output. This idea has been applied to various fields like image classification \cite{Chen.2019}, semantic segmentation \cite{Maag.2020,Rottmann.2020,Fieback.2023}, video instance segmentation \cite{Maag.2021}, and object detection \cite{icaart21,Schubert.2021}. In \cite{matatoken.2025} new input features for the hallucination detection problem have been proposed that outperform classical uncertainty-based features and can be derived from internal LVLM calculations.
\section{Method}
\label{sec:method}
\subsection{LVLM Decoding}
\begin{figure}[t!]
\centering
\begin{subfigure}{0.5\columnwidth}
    \centering
    \includegraphics[width=\columnwidth]{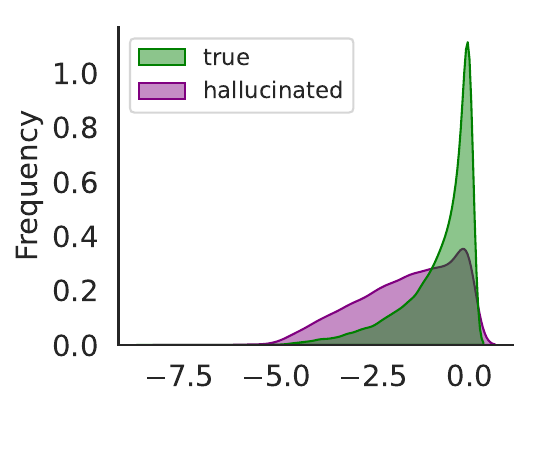}
    \caption{log probability $\log p_\theta$}
    \label{fig_logprob}
\end{subfigure}%
\begin{subfigure}{0.5\columnwidth}
    \centering
    \includegraphics[width=\columnwidth]{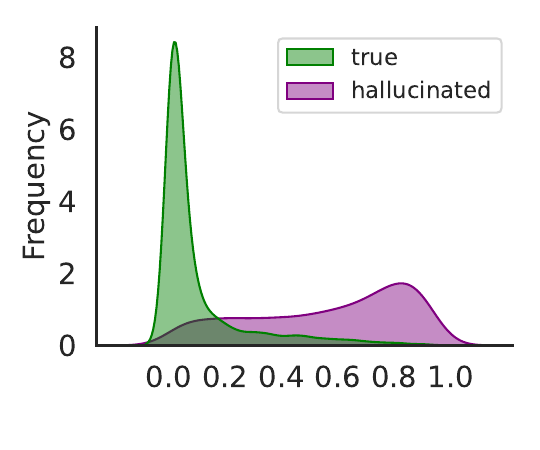}
    \caption{hallucination score $p_{f} $}
    \label{fig_hallucinationscore}
\end{subfigure}%
\caption{Visualization of (a) log probability values and (b) hallucination scores for true and hallucinated tokens.}
\end{figure}

In general, LVLMs generate text responses in an autoregressive way by predicting the probability distribution over the dictionary $\mathcal{V}$ based on the input image $v$, the input query $q$, and the sequence already generated. In the generation step $t$, the next token $y_t \in \mathcal{V}$ is generated by sampling from this distribution. Mathematically, this process can be formulated as
\begin{equation}
    y_t \sim p_\theta ( y_t | v,q,y_{< t}),
\end{equation}
where $\theta$ denote the LVLM parameters and $y_{< t}=(y_{0}, \dots, y_{t-1})$ the generated sequence up to generation step $t-1$. Note that a perfect model should assign high probabilities to true tokens and low probabilities to hallucinations. During this decoding mechanism, hallucinations might be generated when tokens with low probabilities are sampled from $p_\theta ( y_t | v,q,y_{< t})$. However, as we can see in \cref{fig_logprob}, the phenomenon of hallucinations often occurs, as the model assigns high probability values to hallucinated tokens. Our approach corrects this undesired behavior by shifting the final distribution towards true tokens, reducing the probability assigned to hallucinations. 

\subsection{Probabilistic Hallucination Detection} \label{subsec:haldetect}
As we have seen in the previous section, the probability distribution calculated during LVLM decoding does not properly distinguish between true and hallucinated tokens. The idea of meta classification is to learn this classification from hallucination features, which are derived from internal LVLM calculations. By learning from interactions and coherences of these features, the classifier can successfully distinguish between true tokens and hallucinations (see \cref{fig_hallucinationscore}).

To train our classifier, we build on the work of \cite{matatoken.2025} and extend the set of input features to enhance the hallucination detection capabilities. While the features from \cite{matatoken.2025} are based on the last LVLM layer, we integrate further information from the preceding layers into our classifier. This idea is motivated by findings from the LLM literature \cite{chuang2024dola,Azaria.2023} indicating that the middle layers contain information about the reliability of the generated response. To this end, let $N$ denote the number of transformer layers, let $v_0, \dots, v_{u}$ denote the image tokens derived from the vision encoder and alignment module, and $q_0,\dots,q_{w+t}$ the textual tokens representing the input query $q$ and the sequence $y_{< t}$. The concatenated sequence of visual and textual tokens is fed into the LLM and successively processed by each layer $i$ calculating the hidden states $\{ h_{0}^{i}, \dots, h_{u+(w+1)+t}^{i} \}$ with $i = 1,\dots,N$. Finally, the vocabulary head $\phi (\cdot)$ predicts the probability distribution for the next token as
\begin{equation}
    p_\theta ( y_t | v,q,y_{< t}) = \textrm{softmax} \bigl[  \phi (h_{u+(w+1)+t}^{N}) \bigl]_{y_t}, \quad y_t \in \mathcal{V}.
\end{equation}
In order to extract information from the preceding layers, the early exit method \cite{Teerapittayanon.2016,Elbayad2020Depth-Adaptive,Schuster.2022} applies the language head to the hidden states of the earlier layers:
\begin{equation}
\begin{split}
    p_{\theta}^{i} ( y_t | v,q,y_{< t}) & = \textrm{softmax} \bigl[  \phi (h_{u+(w+1)+t}^{i}) \bigl]_{y_t}, \quad \\
    & i \in \{ 1, \dots, N\}.
\end{split}
\end{equation}
For a shorter notation, we write $p_{\theta}^{i}$. Moreover, let $\textrm{Att}^{i,g}_{y_t} (j)$ denote the attention on token $j$ in generation step $t$ for layer $i$ and attention head $g$ with $g = 1,\dots,G$. With this notation, we introduce new features based on intermediate layers:
\begin{itemize}
    \item the negative log-likelihood for all layers
    \begin{equation}
       B^{i} (y_t) = - \log p_{\theta}^{i}, \quad i=1,\dots,N
    \end{equation}
    \item the Kullback–Leibler (KL) \cite{S.KullbackandR.A.Leibler.1951} divergence between the preceding layers and the last layer
    \begin{equation}
    \begin{split}
        K^{i} (y_t) = & \mathrm{KL}(p_{\theta}^{N} || p_{\theta}^{i} ) \coloneq  p_{\theta}^{N} \log \frac{p_{\theta}^{N}}{p_{\theta}^{i}}, \quad \\
        & i=1,\dots,N-1
    \end{split}
    \end{equation}
    \item for each attention head, the entropy of the image attention over the layers
    \begin{equation}
    \begin{split}
        E^{\textrm{layer}}_{v_k,g} (y_t) =  - \frac{1}{N} & \sum_{i=1}^{N} \textrm{Att}_{y_t}^{i,g} (v_k)  \log \textrm{Att}_{y_t}^{i,g} (v_k), \quad \\
        & g = 1,\dots,G
    \end{split}
    \end{equation}
    averaged over the image tokens with
    \begin{equation}
        E^{\textrm{layer}}_{g} (y_t) = \frac{1}{u+1} \sum_{k=0}^{u} E_{v_k,g}^{\textrm{layer}} (y_t)
    \end{equation}
    \item for each layer, the entropy of the image attention over the attention heads
    \begin{equation}
    \begin{split}
        E^{\textrm{head}}_{v_k,i} (y_t) =  - \frac{1}{G} & \sum_{g=1}^{G} \textrm{Att}_{y_t}^{i,g} (v_k)  \log \textrm{Att}_{y_t}^{i,g} (v_k), \quad \\
        & i = 1,\dots,N
    \end{split}
    \end{equation}
    averaged over the image tokens with
    \begin{equation}
        E^{\textrm{head}}_{i} (y_t) = \frac{1}{u+1} \sum_{k=0}^{u} E_{v_k,i}^{\textrm{head}} (y_t).
    \end{equation}
\end{itemize}

We aggregate the features from \cite{matatoken.2025} (see supplementary material A) and our proposed inputs to train the classifier. Let $\mathcal{M}$ denote the set of hallucination features and $m_{y_l}$ the corresponding vector for $y_l$. The meta classifier can be defined as
\begin{align}
\begin{split}
    f:\mathbb{R}^{|\mathcal{M}|} \rightarrow \{0,1\}.
\end{split}
\end{align}
Following \cite{matatoken.2025}, we use the CHAIR evaluation \cite{Rohrbach.2018} to extract true ($z_l = 0$) and hallucinated ($z_l = 1$) tokens from LVLM responses to build our training and validation data with standardized inputs $m_{y_l}$ as
\begin{equation}
    \{ (m_{y_{l}}, z_l) \quad | \quad l = 1,\dots, L \}.
\end{equation}

Once the classifier is trained, we can detect hallucinations during the LVLM generation process by computing the proposed features and applying the classifier afterwards as
\begin{equation}
    f(m_{y_t}) = 
    \begin{cases} 
        1, & p_{f} (y_t | v,q,y_{< t}) \geq \tau \\
        0, & p_{f} (y_t | v,q,y_{< t}) < \tau 
    \end{cases}
\end{equation}
with the estimated probability $p_{f} (y_t | v,q,y_{< t})$ for tokens to be hallucinated, referred to as hallucination scores, and the threshold $\tau$ controlling the precision-recall ratio. Note that the input vector $m_{y_t}$ can be calculated in an automated manner based on internal LVLM calculations only, without any knowledge of the ground truth data.

\subsection{Efficient Contrastive Decoding}
By directly learning hallucinated concepts, we can suppress the generation of hallucinations during the decoding process without an additional LVLM forward pass. In contrast to existing methods, which model the language bias of LVLMs by generating a second output distribution with distorted inputs, we apply a lightweight classifier on the LVLM output to obtain hallucination scores $p_f(y_t | v,q,y_{< t})$, adding only minor computational overhead to the decoding process. At generation step $t$, the contrastive distribution is computed by subtracting the hallucination scores from the log probabilities $\log p_\theta ( y_t | v,q,y_{< t})$ to penalize hallucinations while maintaining high probabilities for true tokens:
\begin{equation}
\begin{split}
    & p_{ecd} ( y_t | v,q,y_{< t}) \\
    & = \textrm{softmax} \bigl[ (1 + \alpha) \log p_\theta ( y_t | v,q,y_{< t}) \\
    & - \alpha \log p_f(y_t | v,q,y_{< t}) \bigl],
\end{split}
\end{equation}
where $\alpha$ controls the magnitude of hallucination correction. Note that for $\alpha = 0$, $p_{ecd}$ is equal to the initial LVLM distribution. Moreover, our proposed efficient contrastive decoding can be integrated into various decoding strategies such as the standard greedy search, beam search \cite{Freitag_2017}, and nucleus sampling \cite{Holtzman2020The}.

\subsection{Adaptive Plausibility Constraint}
We follow the implementation of VCD \cite{Leng_2024_CVPR} and ICD \cite{Wang.2024} and incorporate an adaptive plausibility constraint (APC) \cite{Li.2023} based on the confidence level of the LVLM distribution to maintain high probabilities for semantically trivial tokens. By refining the final contrastive distribution, APC effectively prevents the generation of implausible tokens, and thus, preserves the semantic accuracy of the response. This leads to the final formulation of our proposed decoding strategy:
\begin{equation}
\begin{split}
   & y \sim p_{ecd} ( y_t | v,q,y_{< t}), \quad \\ & \textrm{subject to } y_t \in \mathcal{V}_{\textrm{\scriptsize head}} = \{ y_t \in \mathcal{V} \quad | \quad \\
   & p_\theta ( y_t | v,q,y_{< t}) \geq \beta \max_{\omega} p_\theta ( \omega | v,q,y_{< t}) \}
\end{split}
\end{equation}
with truncation parameter $\beta \in [0,1]$, where $\beta = 1$ implements the standard greedy search algorithm.
\section{Experimental Setup}
\label{sec:experiments}
\subsection{Hallucination Detection}
We evaluate the information content of our proposed input features to learn the differentiation between true and hallucinated answers. First, we sample $5,000$ images from the MSCOCO \cite{Lin.2014} validation set and apply the prompt \textit{"Describe all objects in the image."} to generate training and validation data for the probabilistic classifier. As in \cite{matatoken.2025}, we employ a logistic regression (LR) and gradient boosting (GB) classifier, which have shown superior performance compared to small neural networks in previous studies \cite{Maag.2021}, and use the features from  \cite{matatoken.2025} as our baseline. In detail, we use the LR\footnote{\url{https://scikit-learn.org/stable/modules/generated/sklearn.linear_model.LogisticRegression.html}} classifier with saga solver and the GB\footnote{\url{https://scikit-learn.org/stable/modules/generated/sklearn.ensemble.HistGradientBoostingClassifier.html}} classifier both with $\textrm{max\_iter}=1000$ and scikit-learn version 1.5.2. The detection results are evaluated in terms of accuracy (ACC), area under receiver operator characteristic curve $\textrm{AUROC}$ and area under precision recall curve $\textrm{AUPRC}$ \cite{Davis.2006}. We average our results over ten randomly sampled training-validation splits. The standard deviations can be found the supplementary material A.

\subsection{Datasets and Evaluation Metrics}
\subsubsection{CHAIR} The Caption Hallucination Assessment with Image Relevance (CHAIR) \cite{Rohrbach.2018} metric is widely used in open-ended image captioning tasks and measures the hallucination and coverage rate of LVLMs by checking extracted objects from the generated response against MSCOCO ground-truth labels. CHAIR is defined on the instance level $\textrm{CHAIR}_i$ and sentence level $\textrm{CHAIR}_s$ as
\begin{equation}
\begin{split}
    & \textrm{CHAIR}_i = \frac{| \{ \textrm{hallucinated objects} \} |}{| \{ \textrm{all objects mentioned} \} |} \\
    & \textrm{CHAIR}_s = \frac{| \{ \textrm{captions with hallucinated objects} \} |}{| \{ \textrm{all captions} \} |} \\
   & \textrm{Coverage} = \frac{| \{ \textrm{mentioned objects} \} |}{| \{ \textrm{labeled objects} \} |}.
\end{split}
\end{equation}
For the evaluation of our proposed contrastive decoding method, we sample additional $500$ images from the MSCOCO validation set, which do not overlap with the hallucination detection training data.

\subsubsection{AMBER} An LLM-free Multi-dimensional Benchmark (AMBER) \cite{wang2024amberllmfreemultidimensionalbenchmark}. Since our probabilistic classifier was trained on the MSCOCO dataset, which might lead to biased results in the preceding evaluation, we additionally evaluate our method on the AMBER dataset, which covers a more diverse range of object categories. In detail, AMBER covers $337$ objects compared to $80$ categories for the MSCOCO dataset. The open-ended image captions are again evaluated using $\textrm{CHAIR}_i$, $\textrm{CHAIR}_s$, and $\textrm{Coverage}$ metrics.

\subsubsection{POPE} The Polling-based Object Probing Evaluation (POPE) \cite{Li.2023b} is a discriminative VQA benchmark to assess the quality of LVLMs with respect to object hallucinations. In detail, POPE uses the template \textit{"Is there a \{object\} in the image?"} and applies three different sampling strategies to generate negative prompts, which refer to non-existent objects. The \textit{random} (rand.) sampling chooses the probing objects randomly, \textit{popular} (pop.) samples from high-frequency objects and \textit{adversarial} (adv.) samples among objects, which frequently co-occur with the ground-truth objects. Positive prompts are generated on the basis of ground-truth data. The POPE benchmark covers three datasets, MSCOCO \cite{Lin.2014}, A-OKVQA \cite{Schwenk.2022}, and GQA \cite{Hudson.2019}. For each dataset, POPE samples $500$ images from the validation sets and formulates $6$ probing questions ($3$ positive and $3$ negative prompts) for each image and sampling strategy, yielding a total of $27,000$ question-answer pairs. The results are evaluated in terms of Accuracy, Precision, Recall, and F1 Score.

\subsubsection{MME} The Multimodal LLM Evaluation (MME) benchmark \cite{fu2024mmecomprehensiveevaluationbenchmark} is another discriminative VQA benchmark, which measures perception and cognition abilities of LVLMs on $14$ subtasks comprising $1,193$ images. For each image, there is one positive and one negative question. The evaluation metric is a combined score of the accuracy over all questions and the accuracy+, which is based on each image, that is, both of the two questions need to be answered correctly. Following \cite{Leng_2024_CVPR}, we average the results over five runs. The standard deviations are given in parentheses.

\subsection{Baselines and Implementation Details}
We evaluate our proposed ECD method on three state-of-the-art LVLMs, LLaVA 1.5 \cite{Liu.2023}, InstructBLIP \cite{Dai.2023}, and MiniGPT-4 \cite{zhu2024minigpt}, using nucleus sampling \cite{Holtzman2020The}. For comprehensive parameter settings, we refer to the supplementary material B. We compare our approach against regular decoding (denoted as "regular" in our tables) and the contrastive decoding methods VCD \cite{Leng_2024_CVPR} and ICD \cite{Wang.2024}. Throughout our experiments, we use $\alpha=1$ unless explicitly stated otherwise. All experiments are performed on a single $\textrm{A}100$ GPU.

\section{Results}
\label{sec:results}
\subsection{Hallucination Detection Results}

\begin{table}[t!]
  \centering
  \renewcommand{\arraystretch}{1}
  \setlength{\tabcolsep}{.2em}
  \caption{\textbf{Detection Results.} Hallucination detection results for three LVLMs with the respective hallucination rates given in parentheses. The best results in each block are highlighted.}
  
\begin{center}
   \begin{tabular}{cccc|cc|cc}
        \toprule
        & \multirow{2}*{Set} & \multicolumn{2}{c}{$\textrm{ACC}$ $\uparrow$} & \multicolumn{2}{c}{$\textrm{AUROC}$ $\uparrow$} & \multicolumn{2}{c}{$\textrm{AUPRC}$ $\uparrow$} \\
        && LR & GB & LR & GB & LR & GB \\
        \hline
        LLaVA 1.5 & \cite{matatoken.2025} & $87.05$ & $87.65$ & $89.86$ & $90.82$ & $68.85$ & $71.45$ \\
        \rowcolor{textcolortab!20}\cellcolor{white} ($18.61 \%$) & Ours & $87.93$ & $\textbf{88.27}$ & $91.26$ & $\textbf{91.94}$ & $72.09$ & $\textbf{74.05}$ \\
        \hline
        InstructBLIP & \cite{matatoken.2025} & $91.78$ & $91.94$ & $90.16$ & $90.38$ & $56.71$ & $56.74$ \\
        \rowcolor{textcolortab!20}\cellcolor{white} ($10.13 \%$) & Ours & $\textbf{92.46}$ & $92.26$ & $\textbf{91.79}$ & $91.53$ & $\textbf{61.40}$ & $60.10$ \\
        \hline
        MiniGPT4 & \cite{matatoken.2025} & $89.57$ & $89.70$ & $88.64$ & $88.82$ & $56.64$ & $56.62$ \\
        \rowcolor{textcolortab!20}\cellcolor{white} ($13.05 \%$) & Ours & $90.39$ & $\textbf{90.51}$ & $90.81$ & $\textbf{90.87}$ & $\textbf{62.53}$ & $62.20$ \\
        \bottomrule  
    \end{tabular}
    \end{center}

  \label{tab:results_metatoken} 
   \vspace{-4.5mm}
\end{table}

\begin{table*}[t!]
  \centering
  \renewcommand{\arraystretch}{1}
  \setlength{\tabcolsep}{.35em}
  \caption{\textbf{Discriminative Results on POPE.} Experimental results on the POPE benchmark in terms of accuracy (Acc.), F1 Score and the average inference time per question (time). The best results in each block are highlighted.}
  
\begin{center}
   \begin{tabular}{cccccc|ccc|ccc}
        \toprule
         &  &  & \multicolumn{3}{c}{LLaVA 1.5} & \multicolumn{3}{c}{InstructBLIP} & \multicolumn{3}{c}{MiniGPT4} \\
        & & & time $\downarrow$ & Acc. $\uparrow$ & F1 $\uparrow$ & time $\downarrow$ &  Acc. $\uparrow$ & F1 $\uparrow$ & time $\downarrow$ & Acc. $\uparrow$ & F1 $\uparrow$ \\
        \hline
        
         & & regular & \textbf{0.7} & 87.57 & 87.90 & \textbf{0.5} & 83.90 & 84.01 & \textbf{1.3} & 54.77 & 52.67 \\
         & & VCD & 1.3 & 88.10 & 88.49 & 1.0 & 86.00 & 85.92 & 2.6 & 55.63 & 51.33 \\
         & & ICD & 1.2 & 87.70 & 87.79 & 0.8 & 86.67 & 85.43 & 1.6 & 57.57 & 59.60 \\
         \rowcolor{textcolortab!20}\cellcolor{white} & \cellcolor{white} \multirow{-4}*{\begin{turn}{90} rand. \end{turn}} & Ours & \textbf{0.7} & \textbf{89.00} & \textbf{89.28} & 0.7 & \textbf{89.00} & \textbf{88.69} & 1.4 & \textbf{69.07} & \textbf{72.43} \\
         \cline{3-12}
         & & regular & \textbf{0.6} & 83.50 & 84.57 & \textbf{0.5} & 77.63 & 79.18 & \textbf{1.3} & 48.60 & 49.31 \\
         & & VCD & 1.2 & 85.03 & 85.87 & 1.0 & 78.10 & 79.46 & 2.6 & 50.30 & 48.50 \\
         & & ICD & 1.2 & 85.93 & 86.46 & 0.8 & 79.47 & 78.93 & 1.6 & 52.70 & 57.68 \\
         \rowcolor{textcolortab!20}\cellcolor{white} & \cellcolor{white} \multirow{-4}*{\begin{turn}{90} pop. \end{turn}} & Ours & 0.7 & \textbf{86.97} & \textbf{87.60} & 0.7 & \textbf{82.57} & \textbf{83.23} & 1.4 & \textbf{58.63} & \textbf{66.27} \\
         \cline{3-12}
         & & regular & \textbf{0.6} & 77.93 & 80.38 & \textbf{0.5} & 73.90 & 76.52 & \textbf{1.3} & 47.87 & 48.96 \\
         & & VCD & 1.2 & 78.23 & 80.62 & 1.0 & 75.10 & 77.36 & 2.5 & 48.83 & 47.77 \\
         & & ICD & 1.3 & \textbf{80.07} & 81.53 & 0.8 & 77.77 & 77.77 & 1.7 & 52.63 & 57.90 \\
         \rowcolor{textcolortab!20}{\cellcolor{white}} \multirow{-12}*{\begin{turn}{90} MSCOCO \end{turn}} & \cellcolor{white} \multirow{-4}*{\begin{turn}{90} adv. \end{turn}} & Ours & 0.7 & 79.37 & \textbf{81.58} & 0.7 & \textbf{78.33} & \textbf{79.93} & 1.4 & \textbf{57.27} & \textbf{65.54} \\

         \hline
         & & regular & \textbf{0.6} & 84.77 & 86.22 & \textbf{0.5} & 82.03 & 82.98 & \textbf{1.3} & 49.40 & 47.40 \\
         & & VCD & 1.2 & 84.30 & 85.92 & 1.0 & 82.70 & 83.42 & 2.6 & 52.13 & 48.49 \\
         & & ICD & 1.2 & 85.20 & 86.50 & 0.8 & 85.57 & 84.80 & 1.6 & 54.83 & 58.14 \\
         \rowcolor{textcolortab!20}\cellcolor{white} & \cellcolor{white} \multirow{-4}*{\begin{turn}{90} rand. \end{turn}} & Ours & 0.8 & \textbf{86.50} & \textbf{87.83} & 0.7 & \textbf{87.87} & \textbf{88.16} & 1.4 & \textbf{65.70} & \textbf{70.42} \\
         \cline{3-12}
         & & regular & \textbf{0.6} & 78.17 & 81.47 & \textbf{0.5} & 75.83 & 78.35 & \textbf{1.3} & 46.37 & 47.50 \\
         & & VCD & 1.2 & 78.50 & 81.73 & 1.0 & 76.83 & 78.93 & 2.6 & 47.00 & 44.87 \\
         & & ICD & 1.2 & \textbf{80.20} & 82.69 & 0.8 & 79.33 & 79.39 & 1.6 & 49.10 & 55.13 \\
         \rowcolor{textcolortab!20}\cellcolor{white} & \cellcolor{white} \multirow{-4}*{\begin{turn}{90} pop. \end{turn}} & Ours & 0.8 & 80.03 & \textbf{82.96} & 0.7 & \textbf{80.23} & \textbf{82.13} & 1.4 & \textbf{58.13} & \textbf{66.11} \\
         \cline{3-12}
         & & regular & \textbf{0.6} & 68.80 & 75.25 & \textbf{0.5} & 70.60 & 74.87 & \textbf{1.3} & 43.90 & 46.11 \\
         & & VCD & 1.1 & 69.20 & 75.65 & 1.0 & 70.47 & 74.69 & 2.6 & 45.77 & 45.31 \\
         & & ICD & 1.2 & \textbf{71.63} & \textbf{76.98} & 0.8 & 72.03 & 73.90 & 1.6 & 45.77 & 52.74 \\
         \rowcolor{textcolortab!20}\cellcolor{white} \multirow{-12}*{\begin{turn}{90} A-OKVQA \end{turn}} & \cellcolor{white} \multirow{-4}*{\begin{turn}{90} adv. \end{turn}} & Ours & 0.8 & 69.07 & 75.80 & 0.7 & \textbf{72.40} & \textbf{76.72} & 1.4 & \textbf{53.70} & \textbf{63.57} \\

         \hline
         & & regular & \textbf{0.6} & 84.07 & 85.74 & \textbf{0.6} & 79.97 & 80.95 & \textbf{1.3} & 50.93 & 50.10 \\
         & & VCD & 1.2 & 84.80 & 86.36 & 1.0 & 81.53 & 82.33 & 2.5 & 53.80 & 53.58 \\
         & & ICD & 1.2 & \textbf{86.53} & 87.65 & 0.8 & 83.03 & 81.84 & 1.6 & 55.10 & 59.05 \\
         \rowcolor{textcolortab!20}\cellcolor{white} & \cellcolor{white} \multirow{-4}*{\begin{turn}{90} rand. \end{turn}} & Ours & 0.8 & 86.43 & \textbf{87.78} & 0.7 & \textbf{86.07} & \textbf{86.33} & 1.4 & \textbf{64.10} & \textbf{69.70} \\
         \cline{3-12}
         & & regular & \textbf{0.6} & 74.60 & 78.99 & \textbf{0.6} & 73.57 & 76.34 & \textbf{1.3} & 45.43 & 47.45 \\
         & & VCD & 1.2 & 73.40 & 78.16 & 1.0 & 74.13 & 76.88 & 2.5 & 49.33 & 51.28 \\
         & & ICD & 1.1 & \textbf{75.73} & \textbf{79.72} & 0.8 & 75.93 & 76.22 & 1.6 & 47.83 & 55.20 \\
         \rowcolor{textcolortab!20}\cellcolor{white} & \cellcolor{white} \multirow{-4}*{\begin{turn}{90} pop. \end{turn}} & Ours & 0.8 & 74.63 & 79.14 & 0.7 & \textbf{77.63} & \textbf{79.77} & 1.4 & \textbf{55.50} & \textbf{64.99} \\
         \cline{3-12}
         & & regular & \textbf{0.6} & 69.10 & 75.50 & \textbf{0.6} & 68.73 & 73.18 & \textbf{1.3} & 43.97 & 47.42 \\
         & & VCD & 1.2 & 69.73 & \textbf{75.94} & 1.0 & 70.57 & 74.43 & 2.6 & 46.97 & 49.06 \\
         & & ICD & 1.1 & \textbf{70.03} & 75.89 & 0.8 & 71.47 & 72.84 & 1.6 & 47.23 & 54.91 \\
         \rowcolor{textcolortab!20}\cellcolor{white} \multirow{-12}*{\begin{turn}{90} GQA \end{turn}} & \cellcolor{white} \multirow{-4}*{\begin{turn}{90} adv. \end{turn}} & Ours & 0.8 & 69.10 & 75.89 & 0.7 & \textbf{72.03} & \textbf{76.12} & 1.4 & \textbf{52.80} & \textbf{63.80} \\
        \bottomrule  
    \end{tabular}
    \end{center}

  \label{tab:pope} 
   \vspace{-4.5mm}
\end{table*}

\begin{table*}[t!]
  \centering
  \renewcommand{\arraystretch}{1}
  \setlength{\tabcolsep}{.1em}
  \caption{\textbf{Discriminative Results on MME.} Experimental results on the MME benchmark in terms of Perception and Cognition scores \cite{fu2024mmecomprehensiveevaluationbenchmark}. The best results in each block are highlighted.}
  
\begin{center}
   \begin{tabular}{ccc|cc|cc}
        \toprule
        & \multicolumn{2}{c}{LLaVA 1.5} & \multicolumn{2}{c}{InstructBLIP} & \multicolumn{2}{c}{MiniGPT4} \\
        & Perception $\uparrow$ & Cognition $\uparrow$ & Perception $\uparrow$ & Cognition $\uparrow$ & Perception $\uparrow$ & Cognition $\uparrow$ \\
        \hline
        regular & $1291.1 ^{(\pm 33.1)}$ & $317.4 ^{(\pm 21.0)}$ & $1117.5 ^{(\pm 23.6)}$ & $\textbf{322.1} ^{(\pm 33.3)}$ & $409.9 ^{(\pm 20.2)}$ & $163.6 ^{(\pm 13.0)}$ \\
        VCD & $1288.6 ^{(\pm 33.4)}$ & $338.9 ^{(\pm 16.9)}$ & $1155.2 ^{(\pm 29.9)}$ & $291.1 ^{(\pm 16.4)}$ & $355.9 ^{(\pm 22.2)}$ & $140.1 ^{(\pm 19.5)}$ \\
        ICD & $1314.4 ^{(\pm 27.3)}$ & $318.9 ^{(\pm 22.4)}$ & $\textbf{1258.6} ^{(\pm 19.7)}$ & $295.4 ^{(\pm 35.4)}$ & $\textbf{514.2} ^{(\pm 29.4)}$ & $137.8 ^{(\pm 13.1)}$ \\
        \rowcolor{textcolortab!20} Ours & $\textbf{1400.3} ^{(\pm 13.9)}$ & $\textbf{346.3} ^{(\pm 19.2)}$ & $1105.2 ^{(\pm 17.4)}$ & $282.3 ^{(\pm 13.5)}$ & $502.8 ^{(\pm 17.1)}$ & $\textbf{176.4} ^{(\pm 25.1)}$ \\
        \bottomrule  
    \end{tabular}
    \end{center}

  \label{tab:mme} 
   \vspace{-4.5mm}
\end{table*}

In this section, we evaluate the information content of our proposed input features (see \cref{subsec:haldetect}) for probabilistic hallucination detection. The focus of our evaluation is on the AUPRC values as we observe highly imbalanced datasets, i.e. low hallucination rates. The results for the LR and GB classifier are stated in \cref{tab:results_metatoken}. Our new input features outperform the baseline features in all settings by up to $5.89 pp$ in AUPRC values. While the LR and GB classifier show equal performance on InstructBLIP and MiniGPT4, we observe superiority of the GB model for LLaVa 1.5 with an improvement of $1.96 pp$ in terms of AUPRC. Thus, we employ the GB model in the following experiments.

\subsection{Discriminative Results}\label{subsec:discr_results}

\subsubsection{POPE}

\cref{tab:pope} summarizes our results on the POPE dataset in terms of accuracy (Acc) and F1 Scores. Our proposed ECD method is superior to the baselines in almost all settings improving the F1 Score by up to $23.02 pp$, i.e., $33 \%$. Moreover, ECD always exhibits the best cost-performance trade-off. Note that although the probabilistic hallucination detection was trained on the MSCOCO dataset, our results demonstrate a continuous performance increase across all datasets (MSCOCO, A-OKVQA, and GQA) underlining the ability of the meta classifier to judge hallucinations on new data. Furthermore, we observe consistent performance across all sampling strategies (random, popular, and adversarial) showing that the meta classifier effectively learned hallucinatory concepts beyond the language bias induced by the LVLM training \cite{Rohrbach.2018}. For a detailed analysis of the precision and recall values, we refer to supplementary material B, unveiling the outstanding ability of our method to accurately negate negative prompts, which contain hallucinations. Moreover, the supplementary material contains further experiments with different decoding configurations.

\subsubsection{MME}

The MME benchmark evaluates hallucinations beyond the object level and measures general perception and cognition abilities. \cref{tab:mme} presents our results. For LLaVA 1.5 and MiniGPT4, our method not only improves the perception ability, but also enhances the performance in cognition and reasoning tasks compared to the baseline methods. Moreover, the detailed evaluation of the $14$ subtasks and computational time in supplementary material B demonstrates that ECD outperforms the baselines on most of these individual tasks and demonstrates the superior performance-cost trade-off of ECD. Note that although the averaged ECD perception and cognition scores for InstructBLIP are below the baseline scores, our analysis of the subtasks shows that ECD outperforms the baselines on individual tasks while maintaining low computational costs.

\subsection{Generative Results}

\subsubsection{CHAIR}

\begin{table*}[t!]
  \centering
  \setlength{\tabcolsep}{.1em}
  \caption{\textbf{Generative Results.} Experimental results on the CHAIR benchmark for the open-ended captioning tasks using the MSCOCO and AMBER datasets. The results are stated in terms of average inference time per image caption (time), $\textrm{CHAIR}_i$ ($\textrm{C}_i$), $\textrm{CHAIR}_s$ ($\textrm{C}_s$), and Coverage (Cov.). The best results in each block are highlighted.}
  
\begin{center}
   \begin{tabular}{ccccc|cccc|cccc}
        \toprule
        & \multicolumn{12}{c}{MSCOCO \cite{Lin.2014}} \\
         & \multicolumn{4}{c}{LLaVA 1.5} & \multicolumn{4}{c}{InstructBLIP} & \multicolumn{4}{c}{MiniGPT4} \\
        & time $\downarrow$ & $\textrm{C}_i$ $\downarrow$ & $\textrm{C}_s$ $\downarrow$ & $\textrm{Cov.}$ $\uparrow$ & time $\downarrow$ & $\textrm{C}_i$ $\downarrow$ & $\textrm{C}_s$ $\downarrow$ & $\textrm{Cov.}$ $\uparrow$ & time $\downarrow$ & $\textrm{C}_i$ $\downarrow$ & $\textrm{C}_s$ $\downarrow$ & $\textrm{Cov.}$ $\uparrow$ \\
        \hline
        regular & \textbf{2.7} & 17.86 & 55.00 & 82.14 & \textbf{3.5} & 9.26 & 30.80 & 90.74 & \textbf{10.1} & 11.60 & \textbf{27.47} & 88.40 \\
        VCD & 5.3 & 16.32 & 53.80 & 83.68 & 6.3 & 8.45 & 30.80 & 91.55 & 18.9 & 10.58 & 28.60 & 89.42 \\
        ICD & 4.5 & 14.27 & 45.40 & 85.73 & 5.8 & 10.92 & 37.80 & 89.08 & 15.4 & 10.51 & 28.20 & 89.49 \\
        \rowcolor{textcolortab!20} Ours & 3.6 & \textbf{12.12} & \textbf{43.40} & \textbf{87.88} & 4.6 & \textbf{7.28} & \textbf{26.60} & \textbf{92.72} & 12.8 & \textbf{9.25} & 31.66 & \textbf{90.75} \\
        \hline
        \hline
        
        & \multicolumn{12}{c}{AMBER \cite{wang2024amberllmfreemultidimensionalbenchmark}} \\
         & \multicolumn{4}{c}{LLaVA 1.5} & \multicolumn{4}{c}{InstructBLIP} & \multicolumn{4}{c}{MiniGPT4} \\
        & time $\downarrow$ & $\textrm{C}_i$ $\downarrow$ & $\textrm{C}_s$ $\downarrow$ & $\textrm{Cov.}$ $\uparrow$ & time $\downarrow$ & $\textrm{C}_i$ $\downarrow$ & $\textrm{C}_s$ $\downarrow$ & $\textrm{Cov.}$ $\uparrow$ & time $\downarrow$ & $\textrm{C}_i$ $\downarrow$ & $\textrm{C}_s$ $\downarrow$ & $\textrm{Cov.}$ $\uparrow$ \\
        \hline
        regular & \textbf{2.2} & 9.99 & 45.56 & \textbf{51.97} & \textbf{2.9} & 8.74 & 39.38 & \textbf{51.56} & \textbf{9.4} & 16.68 & 61.78 & 57.71 \\
        VCD & 4.1 & 8.16 & 38.42 & 51.63 & 5.4 & 7.73 & 34.36 & 50.50 & 17.4 & 14.18 & \textbf{53.09} & 58.23 \\
        ICD & 3.5 & 8.46 & 36.87 & 49.98 & 5.3 & 8.11 & 35.71 & 49.02 & 15.3 & 16.83 & 58.11 & 55.35 \\
        \rowcolor{textcolortab!20} Ours & 2.8 & \textbf{7.04} & \textbf{33.20} & 51.21 & 4.2 & \textbf{6.00} & \textbf{26.45} & 50.39 & 12.4 & \textbf{13.14} & 61.58 & \textbf{61.03} \\
        \bottomrule  
    \end{tabular}
    \end{center}

  \label{tab:chair_amber} 
   \vspace{-4.5mm}
\end{table*}

In addition to the discriminative results, we also evaluate our method on the open-ended captioning task. Note that for MiniGPT4, we apply the parameter $\alpha=6$ (see supplementary material B for an ablation study for $\alpha$). The results are summarized in \cref{tab:chair_amber}. ECD distinctly reduces the hallucination rate both at the instance and sentence level, while simultaneously increasing the detailedness of the generated response in terms of Coverage. In all experiments, ECD is superior to the baseline methods VCD and ICD with respect to performance and computational time. Compared to regular decoding, ECD reduces the instance level hallucination rate by up to $5.74 pp$, i.e., $32 \%$ while at the same time increasing the Coverage by $5.74 pp$, i.e., $7 \%$ while maintaining low computational costs.

\subsubsection{AMBER}
Since the ECD meta classifier was trained on the MSCOCO dataset, we investigate the potential of our classifier on new concepts. The results in \cref{tab:chair_amber} underline our findings from the discriminative results (\cref{subsec:discr_results}). Again, ECD successfully suppresses hallucinations due to the classifier's ability to judge hallucinations on new data. More precisely, we reduce the instance level hallucination rate by $2.95 pp$, i.e., $30 \%$ while maintaining the detailedness of the generated response and low computational costs.

\section{Limitations}
\label{sec:limitations}
The focus of our paper is on visual hallucinations of LVLMs, where contextual hallucinations in LLMs might have different origins, which need to be studied to ensure a successful transfer of our method to the unimodal domain. However, note that many hallucination features are specifically designed for the transformer architecture, which can be directly transferred to LLMs, further broadening the impact of this work. Moreover, recent advances in video LVLMs motivate the investigation of temporal hallucinations, a problem we will tackle in future work.
\section{Conclusion}
\label{sec:conclusion}
In this paper, we investigate the power of probabilistic hallucination detection for contrastive decoding. We introduce Efficient Contrastive Decoding (ECD), a lightweight and training-free method, which shifts the output distribution towards accurate responses during decoding by penalizing hallucinations. Extensive experimental results demonstrate the efficacy of our proposed method, which outperforms state-of-the-art methods on various LVLM baselines. Our experiments show that ECD not only mitigates hallucinations, but also enhances the perception capabilities of LVLMs. Moreover, in contrast to existing methods, our lightweight approach is computationally efficient, adding only minor computational overhead to the decoding process.

\subsubsection*{Ethics Considerations}
Our work addresses the hallucination issue in state-of-the-art LVLMs, enhancing the reliability and integrity of LVLMs in real-world scenarios, especially in safety-critical applications such as autonomous driving or medicine. Moreover, our work does not include any personal data, human subjects or sensitive data.

\subsubsection*{Disclaimer} The results, opinions and conclusions expressed in this publication are not necessarily those of Volkswagen Aktiengesellschaft.

\bibliographystyle{splncs04}
\bibliography{haldetect}

\appendix
\label{sec:appendix}
\clearpage
\setcounter{page}{1}
\maketitlesupplementary

\section{Hallucination Detection}
\subsection{Hallucination Features}
The application of meta classification for the hallucination detection problem was first introduced in \cite{matatoken.2025} along with a list of input features to train the classifier. With our notation from the main paper, these features can be defined as:
\begin{itemize}
    \item the \textbf{position} of $y_t$ in the generated response
    \begin{equation} \label{eq:relpos}
        P (y_t) = t
    \end{equation}
    \item the \textbf{absolute occurrence} of token $y_t$ in $y_{< t}$
    \begin{equation} \label{eq:absocc}
       N (y_t) = \sum_{l=0} ^{t} \mathbbm{1}_{\{ y_l = y_t \}}
    \end{equation}
    \item for each attention head, the \textbf{mean attention} of $y_t$ on the image tokens
    \begin{equation} \label{eq:att}
        A^{N,g} (y_t) = \frac{1}{u+1} \sum_{k=0} ^{u} \textrm{Att}_{y_t} (v_k), \quad g = 1,\dots,G
    \end{equation}
    \item the \textbf{log probability}
    \begin{equation} \label{eq:logp}
        L (y_t) = \log p_\theta ( y_t | v,q,y_{< t})
    \end{equation}
    \item the \textbf{cumulated log probability}
    \begin{equation} \label{eq:cumlogp}
        C (y_t) = \sum_{j=0}^{t} \log p_\theta ( y_j | v,q,y_{< t})
    \end{equation}
    \item the sequence \textbf{score} \footnote{\url{https://huggingface.co/docs/transformers/main/en/main_classes/text_generation\#transformers.GenerationMixin.compute_transition_scores}} with length penalty parameter $l_p$
    \begin{equation} \label{eq:score}
        S (y_t) = \frac{1}{t^{l_p}} \sum_{j=0}^{t} \log p_\theta ( y_j | v,q,y_{< t})
    \end{equation}
    \item the \textbf{variance}
    \begin{equation}  \label{eq:var}
        V (y_t) =  \frac{1}{|\mathcal{V}|} \sum_{y \in \mathcal{V}} (\log p_\theta ( y | v,q,y_{< t}) (t) - \mu )^2
    \end{equation} 
    with \quad $\mu = \frac{1}{|\mathcal{V}|} \sum_{y \in \mathcal{V}} \log p_\theta ( y | v,q,y_{< t}) $
    \item the \textbf{entropy} \cite{Shannon.1948}
    \begin{equation} \label{eq:entropy}
        E (y_t) = - \frac{1}{\log |\mathcal{V}|} \sum_{y \in \mathcal{V}} p_\theta ( y | v,q,y_{< t}) \log p_\theta ( y | v,q,y_{< t})
    \end{equation}
    \item the \textbf{variation ratio}
    \begin{equation} \label{eq:varrat}
        R (y_t) = 1 - \max_{y \in \mathcal{V}} p_\theta ( y | v,q,y_{< t})
    \end{equation} 
    \item the \textbf{probability margin}
    \begin{equation} \label{eq:probmargin}
        M (y_t) = R_{o_j} + \max_{y \in \mathcal{V}\setminus \{t_{\mathrm{max}}\}}  p_\theta ( y | v,q,y_{< t})
    \end{equation}
    with $t_{\mathrm{max}} = \max_{y \in \mathcal{V}} p_\theta ( y | v,q,y_{< t})$
    \item the \textbf{probability difference}
    \begin{equation} \label{eq:probdiff}
        D (y_t) = \log p_\theta ( t_{\mathrm{max}} | v,q,y_{< t}) - \log p_\theta ( y_t | v,q,y_{< t}).
    \end{equation}
\end{itemize}

\subsection{Classification Results}
\begin{table*}[h!]
  \centering
  \caption{\textbf{Hallucination Detection Results.} Detection results for the LVLMs LLaVA 1.5 (LV), InstructBLIP (IB), and MiniGPT-4 (MG). The best results in each block are highlighted.}
  
\begin{center}
   \begin{tabular}{cccc|cc|cc}
        \toprule
        & \multirow{2}*{Set} & \multicolumn{2}{c}{$\textrm{ACC}$ $\uparrow$} & \multicolumn{2}{c}{$\textrm{AUROC}$ $\uparrow$} & \multicolumn{2}{c}{$\textrm{AUPRC}$ $\uparrow$} \\
        && LR & GB & LR & GB & LR & GB \\
        \hline
        & \cite{matatoken.2025} & $87.05 ^{(\pm 0.2)}$ & $87.65 ^{(\pm 0.3)}$ & $89.86 ^{(\pm 0.4)}$ & $90.82 ^{(\pm 0.4)}$ & $68.85 ^{(\pm 1.1)}$ & $71.45 ^{(\pm 0.9)}$ \\
        \rowcolor{textcolortab!20}\cellcolor{white} \multirow{-2}*{LV} & Ours & $87.93 ^{(\pm 0.2)}$ & $\textbf{88.27} ^{(\pm 0.3)}$ & $91.26 ^{(\pm 0.3)}$ & $\textbf{91.94} ^{(\pm 0.4)}$ & $72.09 ^{(\pm 1.2)}$ & $\textbf{74.05} ^{(\pm 1.0)}$ \\
        \hline
        & \cite{matatoken.2025} & $91.78 ^{(\pm 0.1)}$ & $91.94 ^{(\pm 0.2)}$ & $90.16 ^{(\pm 1.2)}$ & $90.38 ^{(\pm 1.1)}$ & $56.71 ^{(\pm 5.1)}$ & $56.74 ^{(\pm 4.3)}$ \\
        \rowcolor{textcolortab!20}\cellcolor{white} \multirow{-2}*{IB} & Ours & $\textbf{92.46} ^{(\pm 0.1)}$ & $92.26 ^{(\pm 0.1)}$ & $\textbf{91.79} ^{(\pm 1.0)}$ & $91.53 ^{(\pm 0.8)}$ & $\textbf{61.40} ^{(\pm 4.4)}$ & $60.10 ^{(\pm 4.4)}$ \\
        \hline
        & \cite{matatoken.2025} & $89.57 ^{(\pm 0.3)}$ & $89.70 ^{(\pm 0.3)}$ & $88.64 ^{(\pm 1.8)}$ & $88.82 ^{(\pm 1.4)}$ & $56.64 ^{(\pm 6.5)}$ & $56.62 ^{(\pm 5.9)}$ \\
        \rowcolor{textcolortab!20}\cellcolor{white} \multirow{-2}*{MG} & Ours & $90.39 ^{(\pm 0.2)}$ & $\textbf{90.51} ^{(\pm 0.3)}$ & $90.81 ^{(\pm 1.6)}$ & $\textbf{90.87} ^{(\pm 1.2)}$ & $\textbf{62.53} ^{(\pm 8.8)}$ & $62.20 ^{(\pm 5.5)}$ \\
        \bottomrule  
    \end{tabular}
    \end{center}

  \label{tab:metatoken_full} 
   \vspace{-4.5mm}
\end{table*}

Table \cref{tab:metatoken_full} summarizes our classification results over ten randomly sampled training-validation splits for LLaVA 1.5 \cite{Liu.2023}, InstructBLIP \cite{Dai.2023}, and MiniGPT-4 \cite{zhu2024minigpt} using a logistic regression (LR) and gradient boosting (GB) classifier. The standard deviations are given in parentheses. Ours refers to the combined features from \cite{matatoken.2025} and Section 3.2 from the main paper. Our feature set outperforms the results from \cite{matatoken.2025} under all settings.

\section{Hallucination Mitigation}

\subsection{Implementation Details}

In our experiments, we use the 7B LLM decoder. The detailed configuration settings applied in our experiments are summarized in \cref{tab:lvlm_cnfig}.

\begin{table}[h!]
  \centering
  \setlength{\tabcolsep}{.5em}
  \caption{\textbf{LVLM Generation Configurations.} The generation configurations applied in our experiments for nucleus sampling \cite{Holtzman2020The} and greedy search.}
  
   \begin{tabular}{lcc}
        \toprule
        \multirow{2}*{parameter} & nucleus & greedy  \\
         & sampling & search \\
        \midrule    
        $\textrm{do\_sample}$ & $\textrm{True}$ & $\textrm{False}$ \\
        $\textrm{top\_p}$ & $\textrm{0.9}$ & $\textrm{1}$ \\
        $\textrm{temperature}$ & $\textrm{1}$ & $\textrm{1}$ \\
        $\textrm{num\_beams}$ & $\textrm{1}$ & $\textrm{1}$ \\
        $\textrm{max\_length}$ & $\textrm{256}$ & $\textrm{256}$ \\
        $\textrm{min\_length}$ & $\textrm{1}$ & $\textrm{1}$ \\
        $\textrm{repetition\_penalty}$ & $\textrm{1}$ & $\textrm{1}$ \\
        $\textrm{length\_penalty}$ & $\textrm{1}$ & $\textrm{1}$ \\
        \bottomrule
    \end{tabular}

  \label{tab:lvlm_cnfig} 
   \vspace{-4.5mm}
\end{table}

\subsection{Discriminative Results}

\subsubsection{POPE}

\begin{table*}[h!]
  \centering
  \renewcommand{\arraystretch}{1.0}
  \setlength{\tabcolsep}{.3em}
  \caption{\textbf{Discriminative Results on POPE.} Experimental results on the POPE benchmark \cite{Li.2023b} in terms of accuracy (Acc.), Precision (Prec.), Recall (Rec.) and F1 Score. The best results in each block are highlighted.}
  
\begin{center}
   \begin{tabular}{ccccccc|cccc|cccc}
        \toprule
         &  &  & \multicolumn{4}{c}{LLaVA 1.5} & \multicolumn{4}{c}{InstructBLIP} & \multicolumn{4}{c}{MiniGPT4} \\
        \cline{4-15}
        & & & Acc. $\uparrow$ & Prec. $\uparrow$ & Rec. $\uparrow$ & F1 $\uparrow$ & Acc. $\uparrow$ & Prec. $\uparrow$ & Rec. $\uparrow$ & F1 $\uparrow$ & Acc. $\uparrow$ & Prec. $\uparrow$ & Rec. $\uparrow$ & F1 $\uparrow$ \\
        \hline
         &  & regular & 87.57 & 85.60 & 90.33 & 87.90 & 83.90 & 83.43 & 84.60 & 84.01 & 54.77 & 55.23 & 50.33 & 52.67 \\
         & & VCD & 88.10 & 85.70 & 91.47 & 88.49 & 86.00 & 86.39 & 85.47 & 85.92 & 55.63 & 56.84 & 46.80 & 51.33 \\
         & & ICD & 87.70 & \textbf{87.13} & 88.47 & 87.79 & 86.67 & \textbf{94.14} & 78.20 & 85.43 & 57.57 & 56.87 & 62.60 & 59.60 \\
         \rowcolor{textcolortab!20}\cellcolor{white} & \cellcolor{white} \multirow{-4}*{\begin{turn}{90} rand. \end{turn}} & Ours & \textbf{89.00} & 87.07 & \textbf{91.60} & \textbf{89.28} & \textbf{89.00} & 91.26 & \textbf{86.27} & \textbf{88.69} & \textbf{69.07} & \textbf{65.33} & \textbf{81.27} & \textbf{72.43} \\
         \cline{3-15}
         &  & regular & 83.50 & 79.44 & 90.40 & 84.57 & 77.63 & 74.06 & 85.07 & 79.18 &48.60 & 48.64 & 50.00 & 49.31 \\
         & & VCD & 85.03 & 81.34 & 90.93 & 85.87 & 78.10 & 74.81 & 84.73 & 79.46 & 50.30 & 50.32 & 46.80 & 48.50 \\
         & & ICD & 85.93 & 83.35 & 89.80 & 86.46 & 79.47 & \textbf{81.04} & 76.93 & 78.93 & 52.70 & 52.19 & 64.47 & 57.68 \\
         \rowcolor{textcolortab!20}\cellcolor{white} & \cellcolor{white} \multirow{-4}*{\begin{turn}{90} pop. \end{turn}} & Ours & \textbf{86.97} & \textbf{83.55} & \textbf{92.07} & \textbf{87.60} & \textbf{82.57} & 80.17 & \textbf{86.53} & \textbf{83.23} & \textbf{58.63} & \textbf{55.94} & \textbf{81.27} & \textbf{66.27} \\
         \cline{3-15}
         &  & regular & 77.93 & 72.36 & 90.40 & 80.38 & 73.90 & 69.54 & 85.07 & 76.52 & 47.87 & 47.95 & 50.00 & 48.96 \\
         & & VCD & 78.23 & 72.66 & 90.53 & 80.62 & 75.10 & 70.93 & 85.07 & 77.36 & 48.83 & 48.78 & 46.80 & 47.77 \\
         & & ICD & 80.07 & \textbf{75.95} & 88.00 & 81.53 & 77.77 & \textbf{77.75} & 77.80 & 77.77 & 52.63 & 52.11 & 65.13 & 57.90 \\
         \rowcolor{textcolortab!20}\cellcolor{white} \multirow{-12}*{\begin{turn}{90} MSCOCO \end{turn}} & \cellcolor{white} \multirow{-4}*{\begin{turn}{90} adv. \end{turn}} & Ours & \textbf{79.37} & 73.67 & \textbf{91.40} & \textbf{81.58} & \textbf{78.33} & 74.45 & \textbf{86.27} & \textbf{79.93} & \textbf{57.27} & \textbf{54.91} & \textbf{81.27} & \textbf{65.54} \\
         \hline
         &  & regular & 84.77 & 78.70 & 95.33 & 86.22 & 82.03 & 78.82 & 87.60 & 82.98 & 49.40 & 49.35 & 45.60 & 47.40 \\
         & & VCD & 84.30 & 77.89 & 95.80 & 85.92 & 82.70 & 80.07 & 87.07 & 83.42 & 52.13 & 52.48 & 45.07 & 48.49 \\
         & & ICD & 85.20 & 79.53 & 94.80 & 86.50 & 85.57 & \textbf{89.55} & 80.53 & 84.80 & 54.83 & 54.17 & 62.73 & 58.14 \\
         \rowcolor{textcolortab!20}\cellcolor{white} & \cellcolor{white} \multirow{-4}*{\begin{turn}{90} rand. \end{turn}} & Ours & \textbf{86.50} & \textbf{79.97} & \textbf{97.40} & \textbf{87.83} & \textbf{87.87} & 86.09 & \textbf{90.33} & \textbf{88.16} & \textbf{65.70} & \textbf{61.90} & \textbf{81.67} & \textbf{70.42} \\
         \cline{3-15}
         &  & regular & 78.17 & 70.76 & 96.00 & 81.47 & 75.83 & 70.96 & 87.47 & 78.35 &46.37 & 46.52 & 48.53 & 47.50 \\
         & & VCD & 78.50 & 71.05 & 96.20 & 81.73 & 76.83 & 72.37 & 86.80 & 78.93 & 47.00 & 46.75 & 43.13 & 44.87 \\
         & & ICD & \textbf{80.20} & \textbf{73.45} & 94.60 & 82.69 & 79.33 & \textbf{79.18} & 79.60 & 79.39 & 49.10 & 49.29 & 62.53 & 55.13 \\
         \rowcolor{textcolortab!20}\cellcolor{white} & \cellcolor{white} \multirow{-4}*{\begin{turn}{90} pop. \end{turn}} & Ours & 80.03 & 72.36 & \textbf{97.20} & \textbf{82.96} & \textbf{80.23} & 74.93 & \textbf{90.87} & \textbf{82.13} & \textbf{58.13} & \textbf{55.53} & \textbf{81.67} & \textbf{66.11} \\
         \cline{3-15}
         &  & regular & 68.80 & 62.36 & 94.87 & 75.25 & 70.60 & 65.37 & 87.60 & 74.87 & 43.90 & 44.36 & 48.00 & 46.11 \\
         & & VCD & 69.20 & 62.55 & 95.67 & 75.65 & 70.47 & 65.35 & 87.13 & 74.69 & 45.77 & 45.69 & 44.93 & 45.31 \\
         & & ICD & \textbf{71.63} & \textbf{64.77} & 94.87 & \textbf{76.98} & 72.03 & \textbf{69.27} & 79.20 & 73.90 & 45.77 & 46.73 & 60.53 & 52.74 \\
         \rowcolor{textcolortab!20}\cellcolor{white} \multirow{-12}*{\begin{turn}{90} A-OKVQA \end{turn}} & \cellcolor{white} \multirow{-4}*{\begin{turn}{90} adv. \end{turn}} & Ours & 69.07 & 62.25 & \textbf{96.87} & 75.80 & \textbf{72.40} & 66.34 & \textbf{90.93} & \textbf{76.72} & \textbf{53.70} & \textbf{52.40} & \textbf{80.80} & \textbf{63.57} \\
         \hline
         &  & regular & 84.07 & 77.59 & 95.80 & 85.74 & 79.97 & 77.16 & 85.13 & 80.95 & 50.93 & 50.97 & 49.27 & 50.10 \\
         & & VCD & 84.80 & 78.34 & 96.20 & 86.36 & 81.53 & 78.91 & 86.07 & 82.33 & 53.80 & 53.84 & 53.33 & 53.58 \\
         & & ICD & \textbf{86.53} & \textbf{80.96} & 95.53 & 87.65 & 83.03 & \textbf{88.03} & 76.47 & 81.84 & 55.10 & 54.28 & 64.73 & 59.05 \\
         \rowcolor{textcolortab!20}\cellcolor{white} & \cellcolor{white} \multirow{-4}*{\begin{turn}{90} rand. \end{turn}} & Ours & 86.43 & 79.85 & \textbf{97.47} & \textbf{87.78} & \textbf{86.07} & 84.72 & \textbf{88.00} & \textbf{86.33} & \textbf{64.10} & \textbf{60.29} & \textbf{82.60} & \textbf{69.70} \\
         \cline{3-15}
         &  & regular & 74.60 & 67.36 & 95.47 & 78.99 & 73.57 & 69.10 & 85.27 & 76.34 & 45.43 & 45.76 & 49.27 & 47.45 \\
         & & VCD & 73.40 & 66.30 & 95.20 & 78.16 & 74.13 & 69.50 & 86.00 & 76.88 & 49.33 & 49.38 & 53.33 & 51.28 \\
         & & ICD & \textbf{75.73} & \textbf{68.47} & 95.40 & \textbf{79.72} & 75.93 & \textbf{75.33} & 77.13 & 76.22 & 47.83 & 48.37 & 64.27 & 55.20 \\
         \rowcolor{textcolortab!20}\cellcolor{white} & \cellcolor{white} \multirow{-4}*{\begin{turn}{90} pop. \end{turn}} & Ours & 74.63 & 67.19 & \textbf{96.27} & 79.14 & \textbf{77.63} & 72.81 & \textbf{88.20} & \textbf{79.77} & \textbf{55.50} & \textbf{53.57} & \textbf{82.60} & \textbf{64.99} \\
         \cline{3-15}
         &  & regular & 69.10 & 62.55 & 95.20 & 75.50 & 68.73 & 64.06 & 85.33 & 73.18 & 43.97 & 44.67 & 50.53 & 47.42 \\
         & & VCD & 69.73 & 63.02 & 95.53 & \textbf{75.94} & 70.57 & 65.80 & 85.67 & 74.43 & 46.97 & 47.20 & 51.07 & 49.06 \\
         & & ICD & \textbf{70.03} & \textbf{63.48} & 94.33 & 75.89 & 71.47 & \textbf{69.49} & 76.53 & 72.84 & 47.23 & 47.94 & 64.27 & 54.91 \\
         \rowcolor{textcolortab!20}\cellcolor{white} \multirow{-12}*{\begin{turn}{90} GQA \end{turn}} & \cellcolor{white} \multirow{-4}*{\begin{turn}{90} adv. \end{turn}} & Ours & 69.10 & 62.22 & \textbf{97.27} & 75.89 & \textbf{72.03} & 66.42 & \textbf{89.13} & \textbf{76.12} & \textbf{52.80} & \textbf{51.74} & \textbf{83.20} & \textbf{63.80} \\
        \bottomrule  
    \end{tabular}
    \end{center}

  \label{tab:pope_full} 
\end{table*}

\cref{tab:pope_full} summarizes our results on the POPE dataset introduced in the main paper. In this section, the focus of our evaluation is on the precision and recall values. While our proposed ECD method is superior to the baselines in almost all settings, we especially observe strong improvements in the recall values, showing the outstanding ability of our method to accurately negate negative prompts, which contain hallucinations.

\subsubsection{MME}

\begin{table*}[h!]
  \centering
  \setlength{\tabcolsep}{.3em}
  \caption{\textbf{Discriminative Results on MME.} Experimental results on the 14 subtasks of the MME benchmark in terms of MME scores \cite{fu2024mmecomprehensiveevaluationbenchmark} and computational time. The results are averaged over five runs. The standard deviations are given in parentheses. The best results in each block are highlighted.}
  
\begin{center}
\begin{adjustbox}{angle=90}
   \begin{tabular}{ccccccccccccccccc}
        \toprule
        & & time $\downarrow$ & i $\uparrow$ & ii $\uparrow$ & iii $\uparrow$ & iv $\uparrow$ & v $\uparrow$ & vi $\uparrow$ & vii $\uparrow$ & viii $\uparrow$ & ix $\uparrow$ & x $\uparrow$ & xi $\uparrow$ & xii $\uparrow$ & xiii $\uparrow$ & xiv $\uparrow$ \\
        \hline
        
        & & & 186.00 & 112.33 & 120.33 & 149.00 & \textbf{122.99} & 106.06 & 147.15 & 126.65 & 108.05 & 112.50 & 112.43 & 59.50 & 65.00 & 80.50 \\
        & \multirow{-2}*{reg.} & \multirow{-2}*{\textbf{0.20}} & $^{(\pm 3.74)}$ & $^{(\pm 16.28)}$ & $^{(\pm 6.09)}$ & $^{(\pm 5.12)}$ & $^{(\pm 6.68)}$ & $^{(\pm 5.02)}$ & $^{(\pm 3.74)}$ & $^{(\pm 3.73)}$ & $^{(\pm 4.59)}$ & $^{(\pm 6.52)}$ & $^{(\pm 8.93)}$ & $^{(\pm 9.92)}$ & $^{(\pm 10.12)}$ & $^{(\pm 10.65)}$ \\
        
        & & & 184.67 & 118.67 & 114.67 & 155.67 & 122.52 & 105.82 & 146.80 & 126.95 & 110.80 & 102.00 & \textbf{116.86} & 64.50 & \textbf{85.50} & 72.00 \\
        & \multirow{-2}*{VCD} & \multirow{-2}*{0.36} & $^{(\pm 4.14)}$ & $^{(\pm 11.80)}$ & $^{(\pm 10.51)}$ & $^{(\pm 7.04)}$ & $^{(\pm 1.32)}$ & $^{(\pm 3.72)}$ & $^{(\pm 1.07)}$ & $^{(\pm 3.19)}$ & $^{(\pm 3.38)}$ & $^{(\pm 7.65)}$ & $^{(\pm 5.00)}$ & $^{(\pm 13.73)}$ & $^{(\pm 8.86)}$ & $^{(\pm 15.36)}$ \\
        
        & & & 187.33 & 120.00 & 125.00 & 157.67 & 121.56 & 113.00 & 146.10 & 130.55 & 110.20 & 103.00 & 107.86 & 67.00 & 78.50 & 65.50 \\
        & \multirow{-2}*{ICD} & \multirow{-2}*{0.37} & $^{(\pm 4.16)}$ & $^{(\pm 8.63)}$ & $^{(\pm 2.36)}$ & $^{(\pm 8.54)}$ & $^{(\pm 4.72)}$ & $^{(\pm 3.42)}$ & $^{(\pm 2.31)}$ & $^{(\pm 6.11)}$ & $^{(\pm 3.25)}$ & $^{(\pm 12.59)}$ & $^{(\pm 6.04)}$ & $^{(\pm 7.81)}$ & $^{(\pm 16.25)}$ & $^{(\pm 11.87)}$ \\
        
        \rowcolor{textcolortab!20}\cellcolor{white} & & & \textbf{192.00} & \textbf{140.00} & \textbf{127.00} & \textbf{161.67} & 121.70 & \textbf{121.35} & \textbf{151.60} & \textbf{145.40} & \textbf{117.60} & \textbf{122.00} & 114.29 & \textbf{67.50} & 77.50 & \textbf{87.00} \\
        \rowcolor{textcolortab!20}\cellcolor{white} \multirow{-8}*{LV} & \multirow{-2}*{Ours} & \multirow{-2}*{0.21} & $^{(\pm 2.45)}$ & $^{(\pm 14.64)}$ & $^{(\pm 12.08)}$ & $^{(\pm 6.91)}$ & $^{(\pm 4.94)}$ & $^{(\pm 4.19)}$ & $^{(\pm 2.58)}$ & $^{(\pm 5.70)}$ & $^{(\pm 4.04)}$ & $^{(\pm 9.41)}$ & $^{(\pm 5.91)}$ & $^{(\pm 9.62)}$ & $^{(\pm 8.66)}$ & $^{(\pm 10.17)}$ \\
        \hline

        & & & 175.67 & 68.00 & 62.33 & 112.00 & 127.21 & 110.71 & 145.90 & 138.65 & 100.05 & 77.00 & 96.14 & \textbf{78.00} & \textbf{75.00} & \textbf{73.00} \\
        & \multirow{-2}*{reg.} & \multirow{-2}*{\textbf{0.14}} & $^{(\pm 6.20)}$ & $^{(\pm 5.52)}$ & $^{(\pm 8.67)}$ & $^{(\pm 8.12)}$ & $^{(\pm 2.95)}$ & $^{(\pm 5.78)}$ & $^{(\pm 6.45)}$ & $^{(\pm 2.33)}$ & $^{(\pm 5.98)}$ & $^{(\pm 10.17)}$ & $^{(\pm 9.45)}$ & $^{(\pm 11.34)}$ & $^{(\pm 18.30)}$ & $^{(\pm 14.35)}$ \\
        
        & & & 169.67 & 70.33 & \textbf{69.00} & 121.00 & 133.54 & 105.00 & 147.90 & \textbf{152.10} & 102.65 & \textbf{84.00} & 93.57 & 61.00 & 66.50 & 70.00 \\
        & \multirow{-2}*{VCD} & \multirow{-2}*{0.27} & $^{(\pm 6.62)}$ & $^{(\pm 10.82)}$ & $^{(\pm 9.40)}$ & $^{(\pm 6.72)}$ & $^{(\pm 5.99)}$ & $^{(\pm 2.36)}$ & $^{(\pm 2.96)}$ & $^{(\pm 1.16)}$ & $^{(\pm 1.63)}$ & $^{(\pm 10.79)}$ & $^{(\pm 5.87)}$ & $^{(\pm 13.38)}$ & $^{(\pm 17.51)}$ & $^{(\pm 13.13)}$ \\
        
        & & & 178.00 & \textbf{86.67} & 70.67 & \textbf{154.33} & \textbf{142.65} & \textbf{121.59} & \textbf{159.30} & 142.95 & \textbf{128.40} & 74.00 & 97.86 & 61.50 & 70.00 & 66.00 \\
        & \multirow{-2}*{ICD} & \multirow{-2}*{0.20} & $^{(\pm 2.45)}$ & $^{(\pm 9.19)}$ & $^{(\pm 3.74)}$ & $^{(\pm 2.00)}$ & $^{(\pm 3.89)}$ & $^{(\pm 3.62)}$ & $^{(\pm 1.14)}$ & $^{(\pm 3.30)}$ & $^{(\pm 5.47)}$ & $^{(\pm 5.61)}$ & $^{(\pm 8.26)}$ & $^{(\pm 18.00)}$ & $^{(\pm 23.82)}$ & $^{(\pm 6.44)}$ \\
        
        \rowcolor{textcolortab!20}\cellcolor{white} & & & \textbf{184.00} & 62.00 & 50.67 & 114.33 & 132.24 & 90.59 & 155.35 & 150.45 & 93.05 & 72.50 & \textbf{100.29} & 56.50 & 64.50 & 61.00 \\
        \rowcolor{textcolortab!20}\cellcolor{white} \multirow{-8}*{IB} & \multirow{-2}*{Ours} & \multirow{-2}*{0.17} & $^{(\pm 2.00)}$ & $^{(\pm 6.62)}$ & $^{(\pm 3.27)}$ & $^{(\pm 6.88)}$ & $^{(\pm 3.46)}$ & $^{(\pm 5.67)}$ & $^{(\pm 2.11)}$ & $^{(\pm 0.99)}$ & $^{(\pm 1.51)}$ & $^{(\pm 6.71)}$ & $^{(\pm 6.62)}$ & $^{(\pm 12.10)}$ & $^{(\pm 9.27)}$ & $^{(\pm 4.06)}$ \\
        \hline

        & & & 47.33 & 30.33 & 42.33 & 34.67 & 38.16 & 38.24 & 49.10 & 44.05 & 42.15 & 43.50 & 34.57 & 30.50 & 44.00 & \textbf{54.50} \\
        & \multirow{-2}*{reg.} & \multirow{-2}*{1.03} & $^{(\pm 10.20)}$ & $^{(\pm 7.26)}$ & $^{(\pm 14.44)}$ & $^{(\pm 6.36)}$ & $^{(\pm 3.32)}$ & $^{(\pm 3.87)}$ & $^{(\pm 2.16)}$ & $^{(\pm 3.20)}$ & $^{(\pm 3.20)}$ & $^{(\pm 9.82)}$ & $^{(\pm 7.01)}$ & $^{(\pm 6.20)}$ & $^{(\pm 15.86)}$ & $^{(\pm 21.12)}$ \\
        
        & & & 24.00 & 26.67 & 33.00 & 35.00 & 39.32 & 32.53 & 42.40 & 43.40 & 40.60 & 39.00 & 29.57 & \textbf{31.00} & 34.00 & 45.50 \\
        & \multirow{-2}*{VCD} & \multirow{-2}*{2.10} & $^{(\pm 3.89)}$ & $^{(\pm 9.13)}$ & $^{(\pm 7.56)}$ & $^{(\pm 4.59)}$ & $^{(\pm 3.79)}$ & $^{(\pm 4.44)}$ & $^{(\pm 2.35)}$ & $^{(\pm 2.63)}$ & $^{(\pm 2.42)}$ & $^{(\pm 9.57)}$ & $^{(\pm 3.34)}$ & $^{(\pm 11.58)}$ & $^{(\pm 4.06)}$ & $^{(\pm 11.34)}$ \\
        
        & & & \textbf{68.00} & \textbf{48.67} & \textbf{54.00} & \textbf{60.33} & 37.41 & 47.29 & 57.25 & 48.80 & 39.90 & 52.50 & \textbf{42.29} & 23.50 & 29.50 & 42.50 \\
        & \multirow{-2}*{ICD} & \multirow{-2}*{1.64} & $^{(\pm 9.85)}$ & $^{(\pm 9.15)}$ & $^{(\pm 1.33)}$ & $^{(\pm 8.33)}$ & $^{(\pm 7.85)}$ & $^{(\pm 6.22)}$ & $^{(\pm 3.53)}$ & $^{(\pm 3.29)}$ & $^{(\pm 8.00)}$ & $^{(\pm 6.52)}$ & $^{(\pm 5.60)}$ & $^{(\pm 5.61)}$ & $^{(\pm 13.64)}$ & $^{(\pm 10.84)}$ \\
        
        \rowcolor{textcolortab!20}\cellcolor{white} & & & 47.67 & 18.33 & 47.00 & 39.00 & \textbf{49.66} & \textbf{51.59} & \textbf{65.95} & \textbf{66.45} & \textbf{59.65} & \textbf{57.50} & 39.43 & 27.50 & \textbf{57.50} & 52.00 \\
        \rowcolor{textcolortab!20}\cellcolor{white} \multirow{-8}*{MG} & \multirow{-2}*{Ours} & \multirow{-2}*{\textbf{0.97}} & $^{(\pm 8.60)}$ & $^{(\pm 5.06)}$ & $^{(\pm 6.62)}$ & $^{(\pm 10.98)}$ & $^{(\pm 2.72)}$ & $^{(\pm 2.06)}$ & $^{(\pm 3.86)}$ & $^{(\pm 2.71)}$ & $^{(\pm 2.66)}$ & $^{(\pm 6.52)}$ & $^{(\pm 5.45)}$ & $^{(\pm 9.08)}$ & $^{(\pm 16.36)}$ & $^{(\pm 1.00)}$ \\
        \bottomrule  
    \end{tabular}
\end{adjustbox}
    \end{center}

  \label{tab:mme_full} 
   \vspace{-4.5mm}
\end{table*}

\begin{table*}[t!]
  \centering
  \setlength{\tabcolsep}{.4em}
  \caption{\textbf{Ablation.} Ablation results on the discriminative POPE benchmark \cite{Li.2023b} for nucleus sampling \cite{Holtzman2020The} decoding with $\textrm{top\_p}=1$ and greedy decoding. The best results in each block are highlighted.}
  
\begin{center}
   \begin{tabular}{ccccc|ccc|ccc}
        \toprule
         &  & \multicolumn{3}{c}{LLaVA 1.5} & \multicolumn{3}{c}{InstructBLIP} & \multicolumn{3}{c}{MiniGPT4} \\
        & & time $\downarrow$ & Acc. $\uparrow$ & F1 $\uparrow$ & time $\downarrow$ &  Acc. $\uparrow$ & F1 $\uparrow$ & time $\downarrow$ & Acc. $\uparrow$ & F1 $\uparrow$ \\
        \hline
         & regular & \textbf{0.6} & 82.47 & 83.55 & \textbf{0.5} & 74.53 & 75.97 & \textbf{1.3} & 45.37 & 44.87 \\
         & VCD & 1.2 & 84.13 & 85.13 & 1.0 & 77.53 & 79.13 & 2.6 & 48.33 & 46.40 \\
         & ICD & 1.2 & 85.70 & 86.22 & 0.8 & 79.80 & 79.37 & 1.7 & 52.57 & 57.07 \\
         \rowcolor{textcolortab!20} \cellcolor{white} \multirow{-4}*{\begin{turn}{90} $p=1$  \end{turn}} & Ours & 0.8 & \textbf{86.20} & \textbf{86.85} & 0.6 & \textbf{82.07} & \textbf{82.87} & 1.4 & \textbf{58.07} & \textbf{66.26} \\
         \hline
         & regular & \textbf{0.6} & \textbf{87.70} & 88.21 & \textbf{0.5} & 82.40 & 83.36 & \textbf{1.3} & \textbf{71.20} & \textbf{73.35} \\
         & VCD & 1.2 & \textbf{87.70} & 88.21 & 0.9 & 81.03 & 81.99 & 2.6 & 68.60 & 68.11 \\
         & ICD & 1.2 & 88.10 & \textbf{88.31} & 0.8 & 81.87 & 81.10 & 1.7 & 65.80 & 70.55 \\
         \rowcolor{textcolortab!20} \cellcolor{white} \multirow{-4}*{\begin{turn}{90} greedy  \end{turn}} & Ours & 0.8 & 87.60 & 88.13 & 0.7 & \textbf{83.27} & \textbf{83.83} & 1.4 & 60.17 & 69.98 \\
        \bottomrule  
    \end{tabular}
    \end{center}

  \label{tab:ablation} 
   \vspace{-4.5mm}
\end{table*}

\begin{table*}[h!]
  \centering
  \caption{\textbf{MME Subtasks.} Overview over the MME subtasks from \cref{tab:mme_full}. For each task, MME contains a set of discriminative visual question answering pairs targeting the respective LVLM ability.}
  
\begin{center}
   \begin{tabular}{ccc}
        \toprule
        split & number & name \\
        \hline
        & i & existence  \\
        & ii & count  \\
        & iii & position  \\
        & iv & color  \\
        & v & posters  \\
        & vi & celebrity  \\
        & vii & scene  \\
        & viii & landmark  \\
        & ix & artwork  \\
        \multirow{-10}*{\begin{turn}{90} Perception \end{turn}} & x & OCR  \\
        \hline
        & xi & commonsense reasoning  \\
        & xii & numerical calculation  \\
        & xiii & text translation  \\
        \multirow{-4}*{\begin{turn}{90} Cognition \end{turn}} & xiv & code reasoning  \\
        \bottomrule  
    \end{tabular}
    \end{center}

  \label{tab:mme_names} 
   \vspace{-4.5mm}
\end{table*}

The detailed experimental results on the individual MME subtasks \cite{fu2024mmecomprehensiveevaluationbenchmark} along with the average computational time per image-question pair are stated in \cref{tab:mme_full} for LLaVA 1.5 \cite{Liu.2023} (LV), InstructBLIP \cite{Dai.2023} (IB), and MiniGPT-4 \cite{zhu2024minigpt} (MG). The description of the tasks can be found in \cref{tab:mme_names}. For LLaVA 1.5 and MiniGPT-4, our proposed method outperforms the baselines on most of the individual subtasks. While for InstructBLIP, the ICD \cite{Wang.2024} method shows the best performance in most tasks with an increase of the inference time of $43 \%$, our proposed ECD method outperforms the baselines on individual tasks while maintaining low computational costs.

\subsection{Ablations}

\subsubsection{LVLM Parameter Configurations} We conduct additional experiments with different decoding configurations for the POPE MSCOCO popular setting. The results are summarized in \cref{tab:ablation}. While the results for regular nucleus sampling with $\textrm{top\_p}=1$ are below the $\textrm{top\_p}=0.9$ results from the main paper, all contrastive decoding strategies maintain the performance under the $\textrm{top\_p}=1$ setting. Again, our method outperforms VCD \cite{Leng_2024_CVPR} and ICD \cite{Wang.2024} with respect to performance and computational time. Note that for the greedy search setting, the contrastive decoding methods achieve only a minor performance increase, where for MiniGPT4 the regular decoding performs best.

\subsubsection{ECD Hyperparameter} 
\begin{figure}[t!]
  \centering
  \includegraphics[width=\columnwidth]{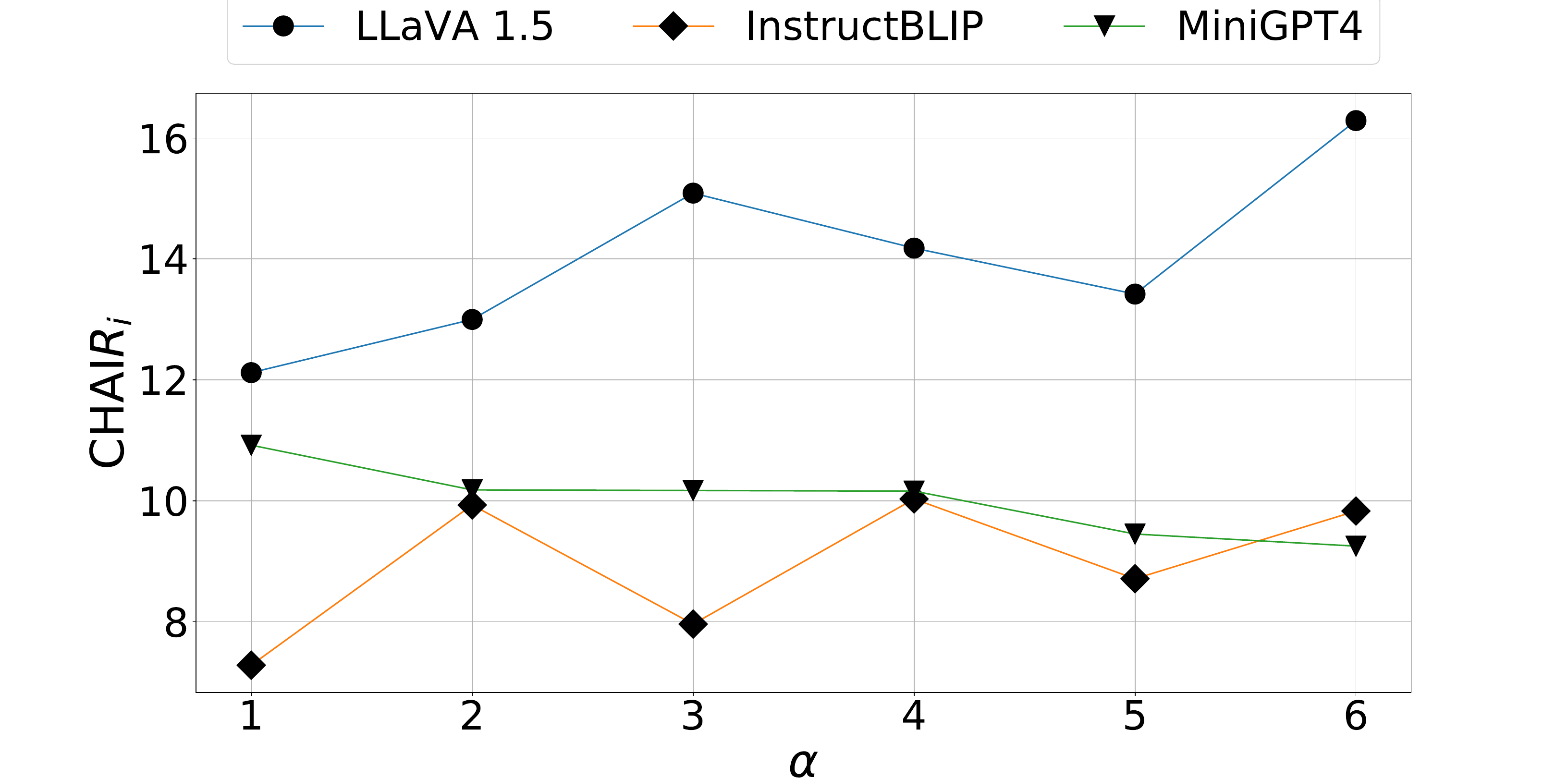}
  \caption{Ablation study for hyperparameter $\alpha$.}\label{fig:alpha}
\end{figure}

Moreover, we investigate the influence of the hyperparameter $\alpha$ in our efficient contrastive decoding (ECD) method in \cref{fig:alpha} on the CHAIR open-ended text generation task. While for LLaVA 1.5 and InstructBLIP the best results are achieved with $\alpha=1$, for MiniGPT4 the best performance is achieved with $\alpha=6$. However, note that the performance changes for MiniPGT4 are small across different $\alpha$ values.

\end{document}